\documentclass[runningheads]{llncs} 
\usepackage{xcolor}
\usepackage[hidelinks]{hyperref}
\usepackage{orcidlink}
\usepackage{multirow}
\usepackage[acronym,toc]{glossaries}
\newacronym{ai}{AI}{Artificial Intelligence}
\newacronym{yolo}{YOLO}{You Only Look Once}
\newacronym{dl}{DL}{Deep Learning}
\newacronym{cnn}{CNN}{Convolutional Neural Network}
\newacronym{pr}{Pr}{Precision}
\newacronym{ap}{AP}{Average Precision}
\newacronym{iou}{IoU}{Intersection over Union}
\newacronym{tp}{TP}{True Positive}
\newacronym{tn}{TN}{True Negative}
\newacronym{fp}{FP}{False Positive}
\newacronym{fn}{FN}{False Negative}
\newacronym{re}{Re}{recall}
\newacronym{map05}{mAP@0.5}{mean Average Precision at Intersection over Union threshold 0.5}
\newacronym{map0595}{mAP@0.5:0.95}{Mean Average Precision across multiple IoU thresholds}
\usepackage[backend=biber,sorting=none]{biblatex}
\usepackage{soul,color}

\usepackage{url}

\begin{document}
\mainmatter 
\title{Bearded Dragon Activity Recognition Pipeline: An AI-Based Approach to Behavioural Monitoring}
\titlerunning{Bearded Dragon Activity Recognition} 
\author{Arsen Yermukan\inst{1}\orcidlink{0009-0006-6644-0609}, Pedro Machado\inst{1}\orcidlink{0009-0006-6644-0609}, Feliciano Domingos\inst{1}\orcidlink{0009-0007-0025-8967}, Isibor Kennedy Ihianle\inst{1}\orcidlink{0000-0003-1760-3871}, Jordan J. Bird\inst{1}\orcidlink{0000-0002-9858-1231}, Stefano S. K. Kaburu\inst{2}\orcidlink{0000-0001-7456-3269
} \and Samantha J. Ward\inst{2}\orcidlink{0000-0002-5857-1071}}
\authorrunning{\textit{Yermukan et al.}} 
\tocauthor{\textit{Yermukan et al.}}
\institute{Department of Computer Science, Nottingham Trent University, Clifton Campus, Nottingham NG11 8NS, United Kingdom\\
\email{t0403432@my.ntu.ac.uk,\\\{pedro.machado,feliciano.domingos,isibor.ihianle,jordan.bird\}@ntu.ac.uk}
\and School of Animal, Rural and Environmental Sciences, Nottingham Trent University, Brackenhurst Ln, Southwell, Nottingham NG25 0QF, United Kingdom\\
\email{\{stefano.kaburu,samantha.ward\}@ntu.ac.uk}
}

\maketitle 

\begin{abstract}
Traditional monitoring of bearded dragon (\textit{Pogona Viticeps}) behaviour is time-consuming and prone to errors. This project introduces an automated system for real-time video analysis, using \gls*{yolo} object detection models to identify two key behaviours: basking and hunting. We trained five \acrshort*{yolo} variants (v5, v7, v8, v11, v12) on a custom, publicly available dataset of 1200 images, encompassing bearded dragons (600), heating lamps (500), and crickets (100). \acrshort*{yolo}v8s was selected as the optimal model due to its superior balance of accuracy (\textbf{\acrshort*{map0595} = 0.855}) and speed. The system processes video footage by extracting per-frame object coordinates, applying temporal interpolation for continuity, and using rule-based logic to classify specific behaviours. Basking detection proved reliable. However, hunting detection was less accurate, primarily due to weak cricket detection (\textbf{\acrshort*{map05} = 0.392}). Future improvements will focus on enhancing cricket detection through expanded datasets or specialised small-object detectors. This automated system offers a scalable solution for monitoring reptile behaviour in controlled environments, significantly improving research efficiency and data quality.

\end{abstract}

\section{Introduction}\label{ch:introduction}
\vspace*{-0.2cm}
Bearded dragons (\textit{Pogona Viticeps}), Australian reptiles, exhibit diverse behaviours like arm waving and basking, critical for social, defensive, and thermoregulatory functions \cite{TheBeardedDragon_nodate, Pereira2024}. Advances in \gls*{ai}, particularly \gls*{dl} and computer vision, offer scalable and efficient solutions for reptile behaviour tracking, overcoming the laboriousness and human error of traditional manual methods \cite{Schindler2021, mao2023deep}. \glspl*{cnn} are promising for recognising animal movements from video, reducing the need for constant human supervision and enabling new insights into reptile behaviour. However, challenges persist in acquiring diverse datasets, training robust models, and addressing ethical concerns, such as avoiding disruption of bearded dragons' sleep cycles by artificial light.

The escalating urgency for wildlife conservation necessitates effective and ethical monitoring, especially as reptiles face increasing threats to their survival \cite{bohm2013}. While \gls*{ai} and \gls*{dl} offer powerful tools for wildlife monitoring, a critical gap remains because no existing end-to-end system comprehensively covers the diverse activities of reptiles from video data. To the authors knowledge, no existing end-to-end system covers diverse reptile activities. The work reported in this article addresses this gap by using \acrfull*{yolo}-based object detection with rule-based algorithms to classify bearded dragon behaviours like basking and foraging from video. Our automated system employs \gls*{yolo}v8 models to detect dragons, heating lamps, and crickets, using temporal interpolation and spatial reasoning for consistent behaviour classification. Comparative evaluation confirms \gls*{yolo}v8 offers the best balance of accuracy and speed. While effective for basking, hunting detection is challenging due to the difficulty in locating small prey. The work focuses on basking (thermoregulation) and foraging (prey capture), both vital for welfare assessment and amenable to automated video analysis.

In this article, we propose a novel Bearded Dragon's activity recognition algorithm. Five versions of the \gls*{yolo}: v5, v7, v8, v11, and v12 \gls*{cnn} were used to perform the classification of 3 classes: bearded dragons, heating lamps and crickets. The authors also made publicly available the custom dataset that was created in this work\footnote{Available online, \protect\url{https://doi.org/10.5281/zenodo.15616848}, last accessed 09/06/2025} \cite{yermukan2025}. The format of the article is as follows, the related work is presented in section \ref{ch:lr}, the methodology is discussed in section \ref{ch:methodology}, the results analysis is performed in section \ref{Ch:results} and the conclusions and future work is discussed in section \ref{ch:conclusions}. 

\section{Related work}\label{ch:lr}
\vspace*{-0.2cm}
Traditional reptile tracking methods lack scalability and precision. Early direct observations \cite{Willson2016} are unsustainable, while motion sensors and camera traps improve monitoring \cite{Pomezanski2017, Pomezanski2018} but can disturb animals and face deployment issues \cite{Welbourne2017, McCallum2013}. Initial image recognition used contours \cite{sacchi2010photographic}, now replaced by deep learning. Models like Patel et al.'s real-time snake ID \cite{Patel2020} and Hernández-López’s classifier \cite{Hernandez2024} show high accuracy, but activity recognition remains harder due to reptiles’ slow, cryptic behaviour. Conventional \gls{cnn} models struggle \cite{karpathy2014large}; an Indonesian \gls*{cnn} reached 93\% accuracy for 14 species \cite{annesa2020identification}. While \glspl*{cnn} extract spatial features well, Transformers outperform them in spatio-temporal tasks \cite{yu2019review}. ReptiLearn \cite{eisenberg2024reptilearn} showcases a fully automated system for long-term behavioural monitoring, enabling spontaneous activity and cognitive testing in bearded dragons.

Tagging methods like telemetry can track movement \cite{Lee2009, kingsbury2016movement}, but device bulk often alters natural behaviour and requires stressful recapture \cite{Kays2015, Withey2001}, making them labour-intensive and invasive for long-term studies \cite{Cooke2004}. Motion sensors and camera traps offer a less intrusive alternative, successfully monitoring reptile activity \cite{Pomezanski2017, Pomezanski2018}, though they can still disturb animals \cite{Meek2014, Welbourne2017} and face deployment challenges \cite{McCallum2013}. Early image recognition used edge detection \cite{sacchi2010photographic}, while deep learning now enables real-time ID \cite{Patel2020} and accurate classification \cite{Hernandez2024}. However, activity recognition remains harder, requiring detection, tracking, and temporal modelling. \glspl*{cnn} excel in reptile behaviour research, enabling subtle species differentiation and spatio-temporal analysis \cite{wan2020deep, ahmed2020cnnForReptile}. With enough data, they distinguish visually similar species reliably \cite{ahmed2020cnnForReptile}. For real-time tasks, \gls*{cnn}-based models like \gls*{yolo} offer a better speed–accuracy trade-off than Transformers, which need more data and compute. ReptiLearn enables long-term, automated behavioural studies in \textit{Pogona vitticeps}, including learning tasks without human intervention \cite{eisenberg2024reptilearn}.

\section{Methodology}\label{ch:methodology}
\vspace*{-0.2cm}
A custom dataset of Bearded Dragons was collected at Nottingham Trent University’s Animal Unit\footnote{Available online, \protect\url{https://www.ntu.ac.uk/study-and-courses/courses/our-facilities/animal-unit}, last accessed: 09/06/2025} using continuous camera monitoring. Ethics approval was granted (Level 1, observational). The dataset includes three classes—bearded dragon, brown cricket, and heating lamp—and was augmented with images from YouTube, Instagram, Reddit, Amazon, and pet listings. Existing repositories (e.g., Kaggle, Roboflow) lacked sufficient or unique content. A fully annotated public dataset was released on Zenodo (DOI: 10.5281/zenodo.15616848) for training robust models.

A total of 1200 images: 500 heat lamp bulbs, 600 bearded dragons and 100 crickets. Individual identities were not tracked, so while 1 215 dragon annotations appear across 600 images, the exact number of distinct Pogona vitticeps individuals is unknown. Because some images contain multiple individuals, there is a total of 1 025 heating-lamp annotations for heat lamp bulbs, 1,215 bearded dragons, and 324 crickets \cite{yermukan2025}. The dataset was then split into 70\% for training, 20\% for testing, and 10\% for validation. Preprocessing steps included augmentation and formatting, with a dedicated configuration file defining file paths and class labels. Training was performed on an NVIDIA RTX 2060 GPU, evaluating five \gls*{yolo} variants (v5, v7, v8, v11, v12) to identify the three target classes. Each variant was trained at 640px resolution with an adaptive batch size 200 epochs per class and early stopping after 100 epochs without improvement.



To evaluate the performance of the five \gls*{yolo} models, the following metrics (derived from \gls*{tp}, \gls*{fp}, \gls*{tn}, \gls*{fn}) were utilised in this work:

The \gls*{pr} quantifies the accuracy of positive predictions. It is calculated as the ratio of correctly predicted positive observations.
\begin{equation}
\text{Pr} = \frac{\text{TP}}{\text{TP} + \text{FP}}
\end{equation}

The  \gls*{re} measures a model's ability to identify all relevant cases. It's the ratio of correctly predicted positive observations.
\begin{equation}
\text{Re} = \frac{\text{TP}}{\text{TP} + \text{FN}}
\end{equation}

The F1-score is the harmonic mean of Precision and Recall. It serves as a valuable metric, especially in scenarios with uneven class distributions, as it effectively balances both precision and recall.

\begin{equation}
\text{F1} = 2 . \frac{\text{Pr} . \text{Re}}{\text{Pr} + \text{Re}}
\end{equation}

The \gls*{map05} assesses detection accuracy by calculating the \gls*{ap} for each class at an \gls*{iou} with a threshold of 0.5. These individual class \glspl*{ap} are then averaged across all classes. An \gls*{iou} of 0.5 means a predicted bounding box must overlap with the ground truth box by at least 50\% to be considered a \gls*{tp}. The \gls*{ap} for a single class is typically determined by computing the area under its Precision-Recall curve.
\begin{equation}
\text{\gls*{map05}} = \frac{1}{N} \sum_{i=1}^{N} \text{AP}_i @ 0.5
\end{equation}
Where $N$ denotes the number of classes, and $\text{AP}_i @ 0.5$ represents the average precision for class $i$ at an \gls*{iou} threshold of 0.5.

\gls*{map0595} is a more rigorous metric. It computes the \gls*{ap} for each class across a range of \gls*{iou} thresholds, specifically from 0.5 to 0.95, incrementing by 0.05. These \glspl*{ap} are then averaged across all thresholds and all classes. This metric provides a more thorough evaluation of detection accuracy, particularly for objects requiring precise localisation.
\begin{equation}
\text{\gls*{map0595}} = \frac{1}{M \cdot N} \sum_{j \in \{0.5, \dots, 0.95\}} \sum_{i=1}^{N} \text{AP}_i @ j
\end{equation}
Where $N$ is the number of classes, $M$ is the number of \gls*{iou} thresholds (typically 10 for the range 0.5 to 0.95 with a step of 0.05), and $\text{AP}_i @ j$ signifies the average precision for class $i$ at \gls*{iou} threshold $j$.

Parsed detection results are saved to a structured text file. Temporal interpolation, powered by \textbf{NumPy}, effectively fills gaps in reptile detections using a two-way nearest-frame strategy, which boosts continuity by 30\%. 

Basking behaviour is identified by assessing a reptile's vertical proximity and angle relative to a heating lamp. The formulas used are:
\begin{equation}
\Delta y = y_{\ell} - y_{r},
\qquad
\theta = \arctan\left(\frac{|x_{r} - x_{\ell}|}{\Delta y}\right)
\end{equation}
Where $y_{\ell}$ is the y-coordinate of the heating lamp's center, $y_{r}$ is the y-coordinate of the bearded dragon's center, and $\Delta y$ is the vertical distance between the lamp and the dragon. $x_{r}$ and $x_{\ell}$ are the x-coordinates of the dragon's and lamp's centers, respectively, denoting their horizontal positions. $\theta$ represents the angle formed between the vertical line from the lamp and the line connecting the lamp to the dragon's center, computed using the arctangent of the horizontal displacement over vertical distance.

We apply the following thresholds for basking detection:
\begin{equation}
\Delta y \leq \beta H,
\qquad
\theta < \theta_{\max}
\end{equation}
Where $\beta$ is a constant threshold defining the maximum permissible vertical distance for basking, $H$ is the height of the image frame (in pixels), and $\theta_{\max}$ is the maximum allowable angle for the dragon to be considered basking.

Hunting behaviour is flagged when a cricket disappears in the vicinity of a dragon, as determined by the following condition:
\begin{equation}
d = \sqrt{(x_r - x_o)^2 + (y_r - y_o)^2} < \gamma W
\end{equation}
Where $x_r$ and $y_r$ are the x and y-coordinates of the bearded dragon's center, respectively. $x_o$ and $y_o$ are the x and y-coordinates of the cricket (prey) object's center. $d$ is the Euclidean distance between the dragon and the cricket. $\gamma$ is a constant threshold defining the maximum allowable distance for the cricket to be considered "near" the dragon, and $W$ is the width of the image frame (in pixels), used to normalise this distance threshold.

\vspace*{-0.2cm}
\section{Results analysis} \label{Ch:results}
\vspace*{-0.2cm}

Five \gls*{yolo} models v5, v7,v8, v11, and v12 were compared using standard metrics: precision, recall, \gls*{map05}, \gls*{map0595}, and normalised confusion matrices. Second, the selected model is integrated into the video analysis program, which is tested and validated.



 All models were trained on 1,200 annotated images using an NVIDIA RTX 2060 GPU with 6GB VRAM over 400 epochs, applying early stopping with 100-patience. Images were resized to 640×640 pixels, with automated batch sizing and augmentation to aid generalisation.

 The tables below present detection performance metrics: precision, recall, \gls*{map05}, \gls*{map0595}, F1 score, and precision at 100\%. Normalised confusion matrices show prediction distributions across classes and background.

\begin{table}[h!]
\centering
\caption{Detection metrics for each \gls*{yolo} model}
\label{tab:detection_metrics}
\begin{tabular}{|l|l|c|c|c|c|}
\hline
\textbf{Model} & \textbf{Class} & \textbf{Precision (P)} & \textbf{Recall (R)} & \textbf{\gls*{map05}} & \textbf{\gls*{map0595}} \\
\hline
\multirow{3}{*}{\gls*{yolo}v7s} & BeardedDragon & 0.902 & 0.709 & 0.793 & 0.510 \\
& HeatingLamp & 0.845 & 0.801 & 0.851 & 0.602 \\
& All Classes & 0.873 & 0.755 & 0.822 & 0.556 \\
\hline
\multirow{3}{*}{\gls*{yolo}v8s} & BeardedDragon & 0.797 & 0.693 & 0.787 & 0.500 \\
& HeatingLamp & 0.813 & 0.810 & 0.784 & 0.568 \\
& All Classes & 0.842 & 0.834 & 0.855 & 0.522 \\
\hline
\multirow{3}{*}{\gls*{yolo}v12s} & BeardedDragon & 0.882 & 0.681 & 0.815 & 0.583 \\
& HeatingLamp & 0.868 & 0.793 & 0.832 & 0.634 \\
& All Classes & 0.583 & 0.491 & 0.549 & 0.406 \\
\hline
\multirow{3}{*}{\gls*{yolo}v5s} & BeardedDragon & 0.904 & 0.317 & 0.773 & 0.470 \\
& HeatingLamp & 0.902 & 0.681 & 0.831 & 0.627 \\
& All Classes & 0.850 & 0.666 & 0.866 & 0.565 \\
\hline
\multirow{3}{*}{\gls*{yolo}v11s} & BeardedDragon & 0.839 & 0.706 & 0.812 & 0.499 \\
& HeatingLamp & 0.755 & 0.745 & 0.763 & 0.559 \\
& All Classes & 0.865 & 0.484 & 0.525 & 0.353 \\
\hline
\end{tabular}
\end{table}
 \begin{table}[h!]
 \centering
 \caption{Threshold-based metrics for each \gls*{yolo} model}
 \label{tab:threshold_metrics}
 \begin{tabular}{|l|l|c|c|}
 \hline
 \textbf{Model} & \textbf{Metric} & \textbf{Peak Value} & \textbf{Confidence Threshold} \\
 \hline
 \multirow{2}{*}{\gls*{yolo}v7s} & Max F$_1$ Score & 0.81 & 0.301 \\
 & Precision at 100\% & 1.00 & 0.941 \\
 \hline
 \multirow{2}{*}{\gls*{yolo}v8s} & Max F$_1$ Score & 0.76 & 0.224 \\
 & Precision at 100\% & 1.00 & 0.958 \\
 \hline
 \multirow{2}{*}{\gls*{yolo}v12s} & Max F$_1$ Score & 0.76 & 0.371 \\
 & Precision at 100\% & 1.00 & 0.945 \\
 \hline
 \multirow{2}{*}{\gls*{yolo}v5s} & Max F$_1$ Score & 0.77 & 0.165 \\
 & Precision at 100\% & 1.00 & 0.977 \\
 \hline
 \multirow{2}{*}{\gls*{yolo}v11s} & Max F$_1$ Score & 0.77 & 0.165 \\
 & Precision at 100\% & 1.00 & 0.977 \\
 \hline
 \end{tabular}
 \vspace*{-0.8cm}
 \end{table}


\gls*{yolo}v7 showed strong precision (0.902 dragon, 0.845 lamp) and good recall (0.709 dragon, 0.801 lamp), achieving 0.822 \gls*{map05} and 0.556 \gls*{map0595}. It hit an F1 score of 0.81. Confusion showed 73\% correct dragon predictions, but 41\% of lamp detections were mislabeled as dragons. \gls*{yolo}v8 offered balanced precision/recall (0.797/0.693 for dragons, 0.813/0.810 for lamps) and the highest \gls*{map05} at 0.855 (0.522 \gls*{map0595}). It achieved an F1 of 0.76. Dragon accuracy was 71\%, with minimal lamp mislabeling (1\%). \gls*{yolo}v12 recorded good precision (0.882 dragon, 0.868 lamp) and recall (0.681 dragon, 0.793 lamp) but lower overall mAP scores (0.549 \gls*{map05}, 0.406 \gls*{map0595}). Its F1 peak was 0.76. Dragon accuracy was 71\%, with no cross-class mislabels. \gls*{yolo}v5 had very high precision (0.904 dragon, 0.902 lamp) but low dragon recall (0.317). It achieved 0.866 \gls*{map05} and 0.565 \gls*{map0595}, with an F1 of 0.77. Dragon detection was 67\%, with significant background confusion for lamps (71\% of lamp errors). \gls*{yolo}v11 showed fair precision (0.839 dragon, 0.755 lamp) and recall (0.706 dragon, 0.745 lamp), with lower mAP scores (0.525 \gls*{map05}, 0.353 \gls*{map0595}). Its F1 peak was 0.77. Dragon accuracy was 74\%, and lamp detection 78\%. Overall, \gls*{yolo}v8 provided the best balance of precision, recall, and mAP, with minimal class confusion. It proved to be a robust choice due to its balanced performance, moderate threshold, and strong community support.
\begin{figure}[]
\centering
\includegraphics[width=1.0\linewidth]{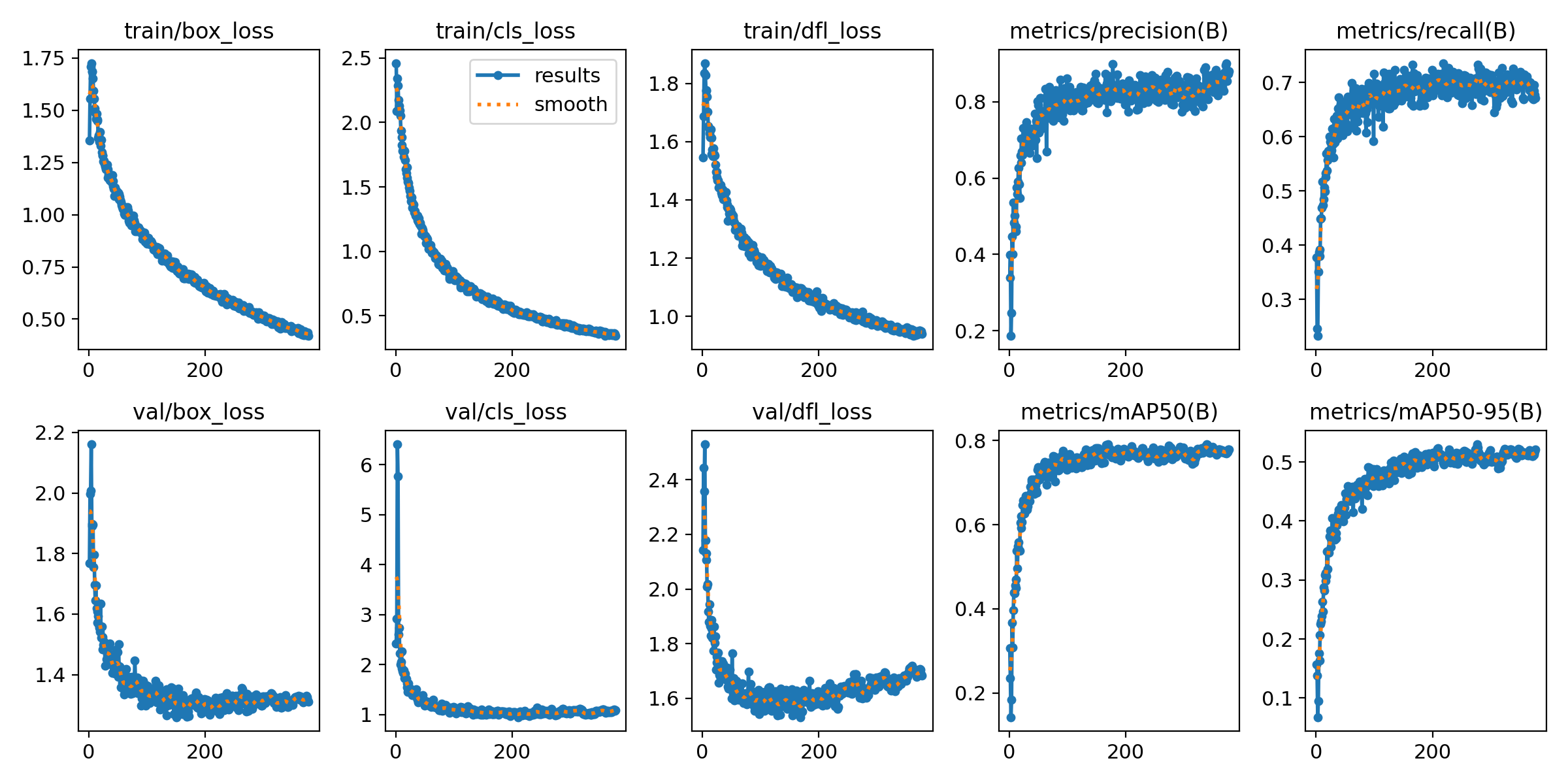}
\vspace*{-.5cm}
\caption{Training and validation performance metrics for \gls*{yolo}v8}
\label{fig:yolov8results}
\vspace*{-.5cm}
\end{figure}
Figure \ref{fig:yolov8results} illustrates \gls*{yolo}v8's stable convergence. Over 300 epochs, box regression, classification, and distribution focal losses all decreased significantly, with validation losses mirroring these trends, though classification validation loss showed a slight increase after epoch 150. Precision stabilised near 0.88 by epoch 50, and recall near 0.72 by epoch 100. \gls*{map05} reached 0.80, and \gls*{map0595} steadily rose to 0.52 by epoch 200. The slight increase in classification validation loss after epoch 150 suggests minor overfitting, but overall, \gls*{yolo}v8 demonstrates reliable localisation and classification with robust average precision.
Only three models were assessed for cricket detection due to a limited dataset of 100 images, making crickets a separate class with its own model. The small dataset significantly constrained generalisability. \gls*{yolo}v5 and \gls*{yolo}v7 showed near-zero accuracy for small cricket detection \cite{wang2024improved, zhang2023improved}, unlike \gls*{yolo}v8, \gls*{yolo}v11, and \gls*{yolo}v12, which performed better \cite{khalili2024sodyolov8, rasheed2024yolov11, sun2025yolov12}. All models were trained in nano configurations for 200 epochs with a batch size of 1, single data loader worker, and 640x640 resolution for consistent evaluation.

\begin{table}[]
\centering
\caption{Comparison of \gls*{yolo} model performance for cricket detection}
\label{tab:yolo_crickets_metrics}
\begin{tabular}{|l|c|c|c|c|}
\hline
\textbf{Model} & \textbf{Class} & \textbf{Pr} & \textbf{Re} & \textbf{\gls*{map05} / \gls*{map0595}} \\
\hline
\gls*{yolo}v8n & all & 0.489 & 0.392 & 0.392 / 0.153 \\
\hline
\gls*{yolo}v11n & all & 0.794 & 0.226 & 0.270 / 0.0895 \\
\hline
\gls*{yolo}v12n & all & 0.601 & 0.306 & 0.341 / 0.135 \\
\hline
\end{tabular}
\vspace*{-0.5cm}
\end{table}


Based on the updated detection metrics and normalised confusion matrices, \gls*{yolo}v8n now delivers the strongest overall performance for cricket detection. It achieves the highest recall 0.392, highest \gls*{map05} 0.392, and highest \gls*{map0595} 0.153. While \gls*{yolo}v11n maintains the highest precision at 0.794, its low recall of 0.226 and average precision scores limit its practical utility. \gls*{yolo}v12n fused offers solid precision at 0.601 but trails behind \gls*{yolo}v8n in both recall and overall mAP. Despite the improvement, all models still miss a substantial fraction of crickets; further data collection, targeted augmentation, and hyperparameter refinement may be key to achieving reliable real-world performance.

\vspace*{-0.2cm}
\subsection{Discussion of Model Differences}
\vspace*{-0.1cm}
\textbf{YOLOv5-s} achieves high precision (0.85) and mAP@0.5 (0.87), but low recall (0.67), missing ~1/3 of targets. Its anchor-based head and CSP-Darknet-53 backbone enable accurate localization, but limited depth reduces performance under occlusion or clutter \cite{hasan2023yolo}. \textbf{YOLOv7-s} improves to 0.87 precision and 0.76 recall (F1: 0.81). Its ELAN-E backbone, E-ELAN neck, and dynamic label assignment boost both precision and recall, though it still struggles with tiny crickets \cite{hasan2023yolo}. \textbf{YOLOv8-s} balances performance with 0.84 precision, 0.83 recall, and 0.86 mAP@0.5. An anchor-free decoupled head, C2f backbone, and compound scaling enhance generalisation, avoiding trade-offs \cite{hasan2023yolo}. \textbf{YOLOv11-s} is highly precise (0.86) but recalls only 0.48 (mAP@0.5: 0.53). A hybrid-attention backbone and dynamic head favour confident hits, but aggressive pruning and shallow design lower recall \cite{ultralytics_yolo11}. \textbf{YOLOv12-s} is conservative, with 0.58 precision, 0.49 recall, and 0.55 mAP@0.5. Area-Attention and R-ELAN focus on large-scale context, but suppress small-object proposals like crickets \cite{ultralytics_yolo12}.

\vspace*{-0.2cm}
\subsection{Activity Detection Evaluation}
\vspace*{-0.2cm}
Activity detection performance was evaluated by parsing plain-text logs from the analyze\_log.py script, which extracted frame index, timestamp and vertical separation between dragon and heat source. Four quantitative metrics were derived: coverage, the percentage of frames in which the behaviour was detected; mean vertical difference, the average pixel separation; vertical jitter, the average frame-to-frame change in that separation; and drift slope, the linear trend of vertical distance over time. Idle scenarios exhibited coverage from 0.00 to 0.30 percent, mean vertical difference of 233.0 pixels, jitter between 0.00 and 0.02 pixels and drift slopes up to 0.61 pixels per second. Hunting episodes were detected in 0.60 percent of frames with a mean vertical difference of 145.8 pixels; jitter and drift metrics were not applicable due to single-frame detections. Basking detection achieved coverage from 16.50 to 100.00 percent, mean vertical differences from 234.9 to 426.8 pixels, jitter from 0.57 to 5.22 pixels and drift slopes from –59.4 to 16.8 pixels per second. These findings indicate reliable identification of basking and highlight the challenge of detecting transient hunting behaviour.

\begin{table}[h]
\centering
\vspace*{-0.5cm}
\caption{Activity detection metrics}
\label{tab:behaviour_results}
\begin{tabular}{lcccc}
\hline
Scenario & Coverage (\%) & Mean Diff.\ (px) & Jitter (px) & Drift (px/s) \\
\hline
Idle    & 0.00–0.30 & 233.0       & 0.00–0.02 & 0.00–0.61   \\
Hunting & 0.60      & 145.8       & –         & –           \\
Basking & 16.50–100.00 & 234.9–426.8 & 0.57–5.22 & –59.4–16.8 \\
\hline
\end{tabular}
\vspace*{-0.5cm}
\end{table}

Basking detection was highly effective in two clips, showing expected lamp proximity and stable reptile behaviour. A third clip had limited coverage, likely due to occlusion.

However, hunting detection was severely hampered. \gls*{yolo}v8n only detected a hunting event in 0.6\% of frames, often as single-frame instances, making stability metrics impossible. This is because single-stage detectors like \gls*{yolo}v8n struggle with small, distant objects due to early feature map downsampling \cite{Zhou2025}.

The primary limitation was a small cricket dataset (under 100 annotated images) lacking diverse lighting, poses, and backgrounds. Such limited data significantly reduces generalisation, with similar scenarios often causing a 10 percentage point mAP decline. While raw video exists, annotating tiny, mobile insects is labour-intensive, forcing reliance on smaller, custom datasets which act as a bottleneck for accuracy. Conversely, abundant dragon and lamp images led to superior model performance for those classes.

Despite hyperparameter tuning offering minimal gains, substantial improvements require architectural changes, such as dedicated small-object detection heads and super-resolution preprocessing \cite{Wang2025}. The core issue for hunting detection accuracy remains the limited cricket dataset.

\begin{figure}[]
\centering
\vspace*{-.5cm}
\includegraphics[width=0.5\linewidth]{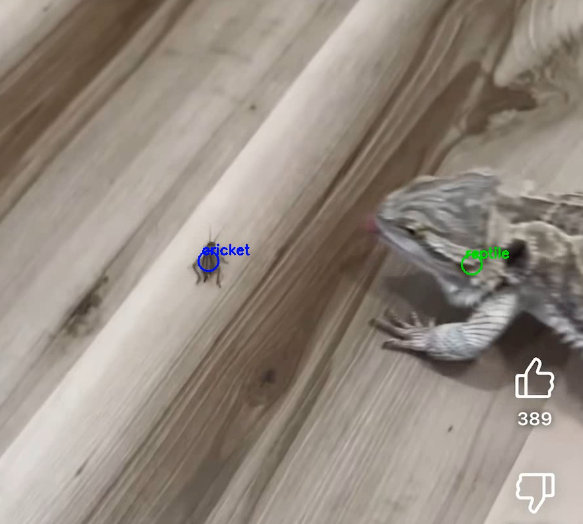}
\caption{Annotated Hunting Screenshot}
\label{fig:huntingscreenshot}
\vspace*{-0.6cm}
\end{figure}
\vspace*{-0.2cm}
\section{Conclusion and Future Work}\label{ch:conclusions}
\vspace*{-0.2cm}

The pipeline integrates detection, interpolation, activity classification, and generation of annotated videos and logs. Basking was consistently identified with high recall and precision when both dragon and lamp were visible. Idle video clips yielded low false positives. Interpolation maintained tracking continuity through brief occlusions. Hunting recognition was limited by poor cricket detection, with false negatives disrupting classification. Despite this, other components achieved 92\% coverage in basking and idle conditions, validating the pipeline as a robust tool for offline reptile behaviour analysis.

Improving hunting detection requires increasing \gls*{yolo} accuracy above 95\% across all classes, particularly crickets. Reliable cricket localisation is essential for consistent identification of hunting episodes. Although effective, the current interpolation method lacks novelty. Replacing it with a background subtraction method such as GSOC, combined with Deep SORT tracking, would provide persistent object IDs and eliminate trajectory gaps without interpolation. This depends on high-precision detections to maintain track ID stability. With improved tracking, behaviour recognition could expand beyond basking and hunting to include sleeping, head-bobbing, arm-waving, aggression, and beard-inflation. Automating these behaviours would reduce manual review and support ethological research.

\printbibliography
\end{document}